\pdfoutput=1
\documentclass[11pt]{article}
\usepackage{subfiles} 
\newtheorem{problem}{Problem}

\usepackage{acl}
\usepackage{multirow}
\usepackage{times}
\usepackage{latexsym}
\usepackage{graphicx}
\usepackage{amsmath}
\DeclareMathOperator*{\argmax}{arg\,max}
\DeclareMathOperator*{\argmin}{arg\,min}
\usepackage[T1]{fontenc}

\usepackage[utf8]{inputenc}

\usepackage{microtype}

\usepackage{booktabs}
\usepackage{multirow}
\usepackage{colortbl}
\usepackage{arydshln}
\usepackage{tabularx}
\usepackage{threeparttable} 

\newcommand{\footlabel}[2]{%
 \addtocounter{footnote}{1}%
 \footnotetext[\thefootnote]{%
  \addtocounter{footnote}{-1}%
  \refstepcounter{footnote}\label{#1}%
  #2%
 }%
 $^{\ref{#1}}$%
}

\title{Description Boosting for Zero-Shot Entity and Relation Classification}
\author{Gabriele Picco \\
  IBM Research Europe \\
  \texttt{ \small gabriele.picco@ibm.com} \\ \And
    Leopold Fuchs\\
  IBM Research Europe \\ 
  \texttt{\small leopold.fuchs@ibm.com} \\ \And 
    Marcos Martinez Galindo\\
  IBM Research Europe \\
  \texttt{\small marcos.martinez.galindo@ibm.com} \\ \AND
    Alberto Purpura\\
  IBM Research Europe \\
  \texttt{ \small alp@ibm.com}\\ \And
    Vanessa Lopez\\
  IBM Research Europe  \\
  \texttt{\small vanlopez@ie.ibm.com} \\ \And
    Hoang Thanh Lam\\
  IBM Research Europe  \\
  \texttt{\small t.l.hoang@ie.ibm.com} 
} 

\begin{document}
\maketitle
\begin{abstract}

 Zero-shot entity and relation classification models leverage available external information of unseen classes - e.g., textual descriptions - to annotate input text data. Thanks to the minimum data requirement, Zero-Shot Learning (ZSL) methods have high value in practice, especially in applications where labeled data is scarce. Even though recent research in ZSL has demonstrated significant results, our analysis reveals that those methods are sensitive to provided textual descriptions of entities (or relations). Even a minor modification of descriptions can lead to a change in the decision boundary between entity classes. In this paper we formally define the problem of identifying effective descriptions for zero shot inference, we propose a strategy for generating variations of an initial description, a heuristic for ranking them and an ensemble method capable of boosting the predictions of zero-shot models through description enhancement. Empirical results on four different entity and relation classification datasets show that our proposed method outperform existing approaches and achieve new SOTA results on these datasets under the ZSL settings. The source code of the proposed solutions and the evaluation framework are open-sourced.~\footnote{\url{https://github.com/IBM/zshot}}

\end{abstract}

\section{Introduction}\label{sec:intro}
Named Entity Recognition (NER) and Relation Extraction (RE) allow for the extraction and categorization of structured data from unstructured text, which in turn enables not only more accurate entity recognition and relationship extraction, but also getting data from several unstructured sources, helping to build knowledge graphs and the semantic web. However, these methods usually rely on labeled data (usually human-annotated data) for a good performance, usually requiring domain experts for data acquisition and labeling, which may incur in high costs. Thus, it is not surprisingly that there is often a lack of labeled data for new domains, limiting the performance of these methods. 

Zero-shot learning (ZSL) is a classification task in machine learning where - at inference time - samples are classified into one of several classes which were not observed during training. In ZSL, the sets of training and test entity classes are disjoint. Therefore, the strategy employed by zero-shot models is to rely on prior general knowledge that could be transferred to unseen instances at inference time. Having a classifier that can generalize to new unseen classes is important for a variety of practical reasons. First, ZSL methods can be used to learn models that are more robust to labeled data shortages and distributional shifts. Moreover, they can be used to extend the reach of models to new domains. 

ZSL approaches in the Natural Language Processing (NLP) domain have seen significant improvements in recent years thanks to the availability of large pre-trained Language Models (LMs). For example, it has been shown that models such as GPT-3 \cite{gpt3}, OPT \cite{zhang2022opt} and FLAN \cite{DBLP:flan} achieve strong performances on many NLP tasks, including translation, question-answering, and cloze tests without any gradient updates or fine-tuning. 




\begin{figure*}[thb]
\center{\includegraphics[width=.8\textwidth]
{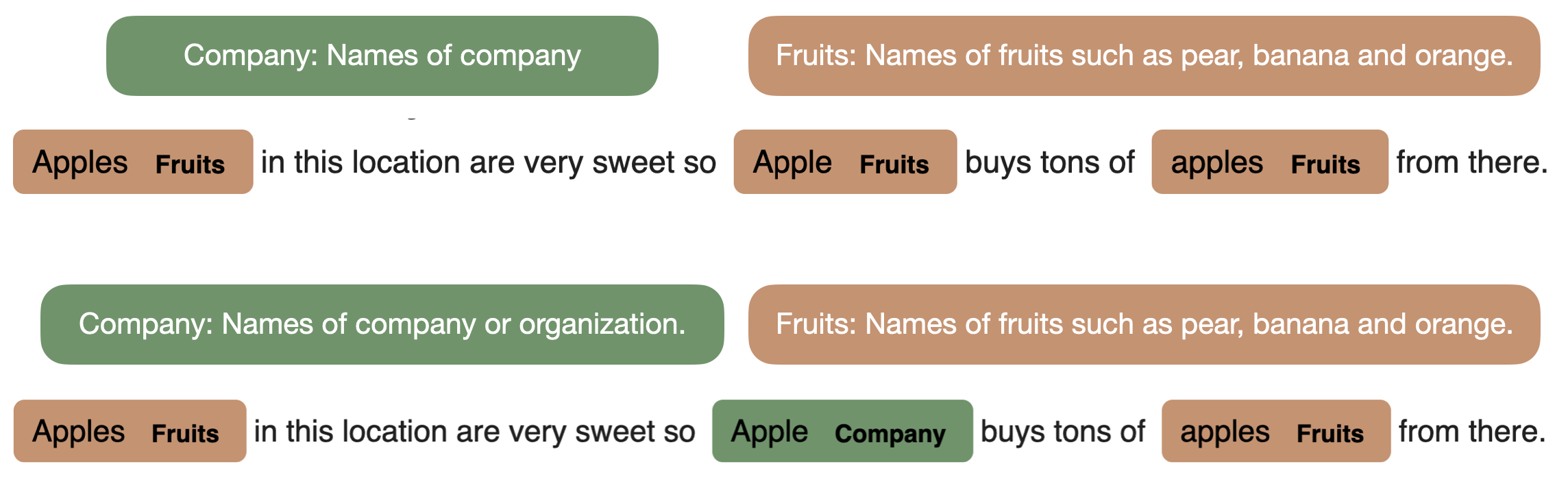}}
\caption{\label{fig:example} A small modification of the \emph{Company} class description results in different entity predictions.}
\end{figure*}

For entity recognition - including classification and linking - and relation classification problems, recent ZSL methods \cite{aly-etal-2021-leveraging, wu2019zero, DBLP:journals/corr/zs-bert} rely on textual descriptions of entities or relations. Descriptions provide the required information about the semantics of entities (or relations), which help the models to identify entity mentions in texts without observing them during training. Works such as \cite{wu2019zero, decao2021autoregressive} and \cite{aly-etal-2021-leveraging} show how effective it is to use textual descriptions to perform entity recognition tasks in the zero-shot context. However, the quality of the descriptions has an impact on how effective the transfer of knowledge from observed to unseen entities \cite{aly-etal-2021-leveraging}. The same mechanism can also be applied in other contexts, such as relation classification \cite{DBLP:journals/corr/zs-bert}. From now on, we refer to entity as both named entities and named relations.

An example of named entity classification with ZSL is demonstrated in Figure \ref{fig:example}. At inference time, a zero-shot model is given short textual descriptions of new entity classes such as \emph{Company} or \emph{Fruits}, it then identifies and annotates mentions of those entity classes in an input sentence. Although state-of-the-art ZSL methods such as SMXM \cite{aly-etal-2021-leveraging} have demonstrated significant results in recent research works, this toy example shows how the quality of the provided descriptions influences the accuracy of these models. For example, in Figure \ref{fig:example} even with a small modification of the \emph{Company} entity class description, the SMXM model changes its entity prediction. In practice, the sensitivity to entity descriptions is problematic because, for non-expert users, it is not a trivial task to choose a proper description for black-box zero-shot models, in particular in an unfamiliar domain. 

In this paper, we study different methods for improving the descriptions in an unsupervised way. Specifically, we propose UDEBO (for Unsupervised DEscription BOosting), the first unsupervised method capable of automatically modifying/generating descriptions to improve entity predictions in the zero-shot settings. We present several strategies to alter descriptions, such as using a generative model, paraphrasing, and summarization combined with description ranking/ensemble methods to reduce model uncertainty and increase overall performance. 
We empirically evaluate the performance of UDEBO on 4 existing standard zero-shot datasets, spanning two tasks: (i) named entity classification and (ii) relation classification. 

Our results show that for the zero-shot entity classification tasks, UDEBO improved the results of state-of-the-art models by 7 and 1.3 percentage points in terms of Macro F1 Score in the OntoNotes and MedMentions datasets, respectively. 
For what concerns relation classification, we achieve a performance improvement of 6 and 3 percentage points (Macro F1 Score) on the FewRel and WikiZS datasets over our baseline models, respectively.

We organize the paper as follows. In Section \ref{ss:problem definition} we formally define the problem we aim to solve in this paper, i.e. how to enhance entity or relation descriptions to improve the performance of zero-shot models.
In Section \ref{s:method} we describe the proposed approaches for description boosting while in Section \ref{sec:experimental_results} we describe our experimental setup and results. We provide a literature review, a discussion about Large Language Models (LLMs) and draw the conclusions of our work in Sections \ref{sec:related}, \ref{sec:discussion} and \ref{sec:conclusions}, respectively.

\section{Problem Definition}
\label{ss:problem definition}



Given a set of entity classes $E$ of interest with their textual descriptions $D$ and a corpus of sentences $S$ to annotate as input, we can define the problem of description enhancement as follows:
 \begin{problem}[Description enhancement]
  Denote $\phi(D, S)$ as the function estimating the accuracy of ZSL models when using a given entity description $D$ for annotating an input text corpus $S$. Our goal is to generate a set of descriptions $D^*$ such that:
 \begin{eqnarray}
  D^* = \argmax_{D} \phi(D, S) 
 \end{eqnarray}
 \label{prob:acc}
 \end{problem}
 
In Section \ref{sec:desc_alter_strategies} we describe different strategies to generate new entity descriptions $D'$ for the input set $E$, intending to improve the accuracy of the predictions by the ZSL models over that corpus. If the labeled data is known, it is possible to select the best descriptions via a brute force search across different description reformulations by measuring the accuracy as a function of $D$ and $S$. However, given the absence of labeled data in the zero-shot context, an unsupervised approach is needed for ranking the descriptions $D$ that yield the highest classification accuracy. In Section \ref{sec:etropy_heuristic} and Section \ref{sec:ensemble}, we will discuss methods for ranking or combining predictions from different description variations to achieve better results.

\section{Methods}
\label{s:method}

The UDEBO approach comprises 2 steps. First, the descriptions are generated or improved. Finally, the descriptions are ranked in order to select the best ones. As an optional step, we analyze the ensembling of descriptions for boosting performance.

\subsection{Generating description variations}\label{sec:desc_alter_strategies}

Improving the completeness or clarity of entity descriptions is a complicated problem without a formal definition of an objective function, as there is a large space of candidates to explore.
To enhance entity descriptions, in a more controlled way, we propose the following strategies.

\paragraph{Extension with pre-trained LMs.} We propose to use large pre-trained LMs for generating text using the given description as context. Large LMs, as shown in \cite{petroni2019language}, capture linguistic and relational knowledge that can be extracted trough generation to extend a given description. In Section \ref{sec:experimental_results} we analyse the use of GPT-2 \cite{radford2019language} for generating descriptions variations.
 
\paragraph{Extension with a fine-tuned LM.} We fine-tune a LM for description generation and expansion. The LM is fine-tuned on a large dataset containing about 5.3 million Wikidata instances, including the name and the first few sentences of the respective articles. The model is fine-tuned on extending a truncated sub-string of the textual description, using a sequence to sequence objective. In Section \ref{sec:experimental_results} we analyse the use of a T5 large \cite{2020t5} fine-tuned model for generating descriptions variations.
 
\paragraph{Summarization.} Text summarization can be used to generate a concise description with less noise compared to the original one. In the experimental results we analyse the effect of using a BERT2BERT (small) \cite{turc2019} model fine-tuned on CNN/Dailymail for text summarization to enhance entities descriptions.

\paragraph{Paraphrasing.} Paraphrasing a description can simplify its linguistic form, using more common and general terms. In the experimental results we analyse the effect of using a Pegasus \cite{zhang2019pegasus} model fine-tuned for paraphrasing. 

\subsection{Description ranking via entropy} \label{sec:etropy_heuristic}

To rank a description for an entity, we propose to use a zero-shot model to first compute the probabilities of classes for each mention in the input text with a candidate description. We then compute the information entropy $H$ from this input. In information theory, entropy is the average level of "information" or "uncertainty" inherent to a variable's possible outcomes. Our assumption is that the lower the entropy is, the higher the confidence of the prediction will be, so Problem \ref{prob:acc} can be reformulated as:
\begin{equation}
 D^* = \argmin_{D} H(D, S) 
\end{equation}
Where $H$ is the entropy of a zero-shot model for a corpus $S$, using the description $D$ to accomplish a certain classification task. This way we can rank different candidate descriptions and choose the best one without requiring any labeled data, which is ideal for the zero-shot setting.

\subsection{Boosting performances with descriptions variations ensembling}
\label{sec:ensemble}
Besides description ranking via entropy, we propose an ensemble method that combines predictions from multiple pipelines executed with different entity descriptions. The main idea behind this approach is to leverage the complementary information provided by the different definitions to make a more accurate prediction, reducing the variance and bias of an individual pipeline. 
Furthermore, using the methods described in section \ref{sec:desc_alter_strategies}, the descriptions variations can provide additional information useful for correctly discriminating between unseen classes.
\begin{table}[tb]
 \centering
 \resizebox{\columnwidth}{!}{
 \begin{tabular}{lccc}
  \hline
  Dataset & Split & Instances & \begin{tabular}{@{}c@{}} Entities / Relations\end{tabular} \\
  \hline
  \multirow{ 3}{*}{MedMentionsZS} &  train &  26770 & 11 \\ \cline{2-4}
  & val &  1289 & 5 \\ \cline{2-4}
  &   test &  1048 & 5 \\ 
  \cline{2-4}
  \hline
  \multirow{ 3}{*}{OntoNotesZS} & train &  41475 & 4 \\ \cline{2-4}
  
  & val &  1358 & 4 \\ \cline{2-4}
  
  &  test &  426 & 3 \\
  \hline
  \multirow{2}{*}{Fewrel} & train &  44800 &   64 \\
  \cline{2-4}
   & test &  11200 &   16 \\
  \cline{2-4}
  \hline
  \multirow{3}{*}{WikiZS} & train &  70952 &   83 \\
  \cline{2-4}
  & val & 12982 &   15 \\
  \cline{2-4}
  & test &  9494 &   15 \\
  \hline  
 \end{tabular}}
 \caption{Number of sentences and entities for each split of the considered datasets.}
 \label{tab:rows_entities_relations}
\end{table}

\paragraph{Entity description ensemble.} Given a sentence, for each span $s$ and an entity label $e \in E$, denote $v(s, e)$ as the number of pipelines that predict $s$ or a sub-sequence of $s$ with entity label $e$. For instance, given a span $s =$ \emph{London Bridge}, assume that among ten pipelines, four pipelines predict the label of $s$ as $e_1=$ Facility, the other four pipelines predict the label of \emph{London} as $e_2=$ Location and the rest of the pipelines predict \emph{Bridge} as Facility. Therefore, the accumulated number of votes for the span \emph{London Bridge} are $v(s, e_1) = 6$ and $v(s, e_2) = 4$. Considering the majority of the votes, the final predicted label for the span \emph{London Bridge} is Facility. Once the span \emph{London Bridge} has been assigned a label, all of its sub-spans become redundant and thus are removed from consideration.


\section{Experiments and Results}\label{sec:experimental_results}
This section discusses experimental settings, baseline methods, and empirical results for both entity and relation classification tasks.

\begin{table*}[htb]
\begin{tabular}{llcccccc}
\hline
{Datasets}   & {Methods} & {Precision} & {Recall} & {Micro F1} & {Macro F1} & {Accuracy} \\
\hline
\multirow{2}{*}{OntoNotesZS} & SMXM    & 20.96    & 48.15 & 30.76    & 29.12    & 86.36   \\
        & SMXM (Pre-trained)  & 24.05    & \textbf{51.40} & 32.77    & 32.78    & 87.69  \\
        & SMXM (Finetuned)  & 17.97    & 42.21 & 25.21    & 23.90    & 85.76 \\
        & SMXM (Summarization) & 18.93    & 35.45 & 24.68    & 19.47    & 85.93   \\
        & SMXM (Paraphrased)  & 18.49    & 40.90 & 25.46    & 23.41    & 85.14   \\
        & SMXM (Combined)  & 18.86    & 42.58 & 26.15    & 23.74    & 84.83   \\
        & UDEBO   & \textbf{31.14}  & 46.51   & \textbf{36.78} & \textbf{36.15} & \textbf{88.29} \\
\hline
\multirow{2}{*}{MedMentionsZS} & SMXM    & 16.79    & \textbf{40.55} & 20.38    & 21.70    & 83.05   \\
        & SMXM (Pre-trained)  & 13.25    & 37.98 & 19.64    & 18.26    & 81.88   \\
        & SMXM (Finetuned)  & 13.67    & 36.05 & 19.82    & 19.13    & 83.18  \\
        & SMXM (Summarization) & 10.96    & 26.68 & 15.37    & 17.92    & 83.02  \\
        & SMXM (Paraphrased)  & 14.77    & 26.51 & 18.97    & 19.41    & 86.74  \\
        & SMXM (Combined)  & 12.80    & 37.15 & 19.04    & 17.92    & 81.63  \\
        & UDEBO   & \textbf{19.51}  & 32.73   & \textbf{23.86} & \textbf{22.97} & \textbf{85.70} \\
\hline
\end{tabular}
\caption{UDEBO, i.e. the ensemble of predictions with description variations, compared to the SMXM baseline.}
\label{tab:ensemble}
\end{table*}

\subsection{Datasets and experimental settings}
\label{sec:datasets}
We use two different settings: one for the Entity Classification (EC) task and one for the Relation Classification (RC) one. 

\paragraph{Entity Classification setting.}
We use the pre-trained SMXM model \cite{aly-etal-2021-leveraging} with the checkpoints available in the official GitHub repository.~\footlabel{footnote:smxm-repo}{\url{https://github.com/Raldir/Zero-shot-NERC/}} We refer the reader to the original paper \cite{aly-etal-2021-leveraging} to see the details of the implementation, the training parameters, and the datasets used for fine-tuning the model. There are two different checkpoints, one for each one of the datasets used, OntoNotes \cite{pradhan-etal-2013-towards} and MedMentions \cite{mohan2019medmentions}. Both datasets have been processed as in the respective official GitHub repositories. Table \ref{tab:rows_entities_relations} shows the number of rows and the entities of each dataset. Note that the number of rows reported in Table \ref{tab:rows_entities_relations} refers to the zero-shot version of the dataset, containing only sentences with entities. See Appendix \ref{appendix:datasets} for more information on this process and the datasets. The results reported are all based on the \emph{test} split of the datasets.

\paragraph{Relation Classification setting.}
For RC, we use ZS-BERT~\footlabel{footnote:zsbert-repo}{\url{https://github.com/dinobby/ZS-BERT}} \cite{DBLP:journals/corr/zs-bert}, a multitask learning model, based on BERT, to directly predict unseen relations. We trained our checkpoint using the official implementation of the model and following the steps of the official repository.~\footref{footnote:zsbert-repo} The datasets we use are FewRel \cite{han-etal-2018-fewrel} and WikiZS \cite{sorokin2017context}. 
The results reported are all based on the \emph{test} split of the datasets.

\paragraph{Description alteration settings.} The language models used for the description alteration strategies: summarization, paraphrasing and pre-trained were obtained from the checkpoints available on Huggingface, while for the latter strategy we have fine-tuned a pre-trained T5-large model. We report detailed hyper-parameters of description alteration methods in section \ref{appendix:hp} of the appendix.

\begin{table*}[]
\begin{tabular}{llcccccc}
\hline
{Datasets}   & {Methods} & {Precision} & {Recall} & {Micro F1} & {Macro F1} & {Accuracy} \\
\hline
\multirow{7}{*}{Fewrel \hphantom{FewrelD} } & ZS-BERT     & 25.08  & 21.59 & 21.59   & 17.89   & 21.59   \\
        & ZS-BERT (Pre-trained)  & 18.25    & 25.29 & 25.29    & 19.10    & 25.29   \\
        & ZS-BERT (Finetuned)  & 19.39    & 16.09 & 16.09    & 14.59    & 16.09 \\
        & ZS-BERT (Summarization) & 19.83    & 19.81 & 19.81    & 15.21    & 19.81   \\
        & ZS-BERT (Paraphrased)  & 25.89    & 21.76 & 21.76    & 19.90    & 21.76  \\
        & ZS-BERT (Combined)  & 17.09    & 16.53 & 16.53    & 16.53    & 16.53    \\
        & UDEBO   & \textbf{28.38} & \textbf{25.68}  & \textbf{25.68} & \textbf{22.12} & \textbf{25.68} \\ \hline
\multirow{7}{*}{WikiZS} & ZS-BERT & 34.18 & 33.90 & 37.14 & 30.97 & 37.14 \\
      & ZS-BERT (Pre-trained)& 14.73 & 15.80 & 14.29 & 11.72 & 14.29 \\
      & ZS-BERT (Finetuned)& 16.23 & 16.26 & 16.62 & 13.65 & 16.62 \\
      & ZS-BERT (Summarization)& 19.07 & 19.57 & 19.62 & 16.87 & 19.62 \\
      & ZS-BERT (Paraphrased) & 25.50 & 27.60 & 27.60 & 24.56 & 27.60 \\
      & ZS-BERT (Combined) & 17.34 & 19.62 & 18.43 & 16.27 & 18.43 \\
      & UDEBO & \textbf{34.79} & \textbf{37.11} & \textbf{40.17} & \textbf{34.25} & \textbf{40.17} \\
\hline
\end{tabular}
\caption{
UDEBO, i.e. the ensemble of predictions with description variations, compared to the ZS-BERT baseline.}
\label{tab:ensemble_relations}
\end{table*}

\subsection{Empirical results}
\label{ss:results}
This section discusses the results of entity classification using methods for description enhancement. 
\subsubsection{Entity classification}

Table \ref{tab:ensemble} shows the results of the ensemble method (UDEBO) with ten descriptions generated by each of the description enhancing strategies, including pre-trained, finetuning, summarization and paraphrasing. For each enhancing strategy, we report the results when the descriptions with the lowest entropy are chosen for each class. The \emph{Combined} strategy shows the results with the lowest entropy among all description-enhancing strategies. 

We can see that the ensemble method (UDEBO) outperforms the SMXM baseline using the original descriptions provided on the OntoNotesZS dataset with a significant margin of 7 percentage points in terms of Macro F1 Score. On the MedMentionZS dataset, the improvement is 1.3 percentage points on the same reference performance measure (Macro F1 Score). Description ranking based on entropy works well with the pre-trained strategy on OntoNotesZS. However, the entropy does not seem to be a reliable score of model uncertainty on the MedMentionsZS dataset. Finding an alternative uncertainty score to entropy could be considered as future work. 
Overall, these results confirm our hypothesis - discussed in Section \ref{sec:intro} - that zero-shot methods are sensitive to provided descriptions and that an ensemble of description enhancement methods is needed to obtain more robust results. 
\begin{figure}[bth]
 \centering
 \includegraphics[width=\columnwidth]{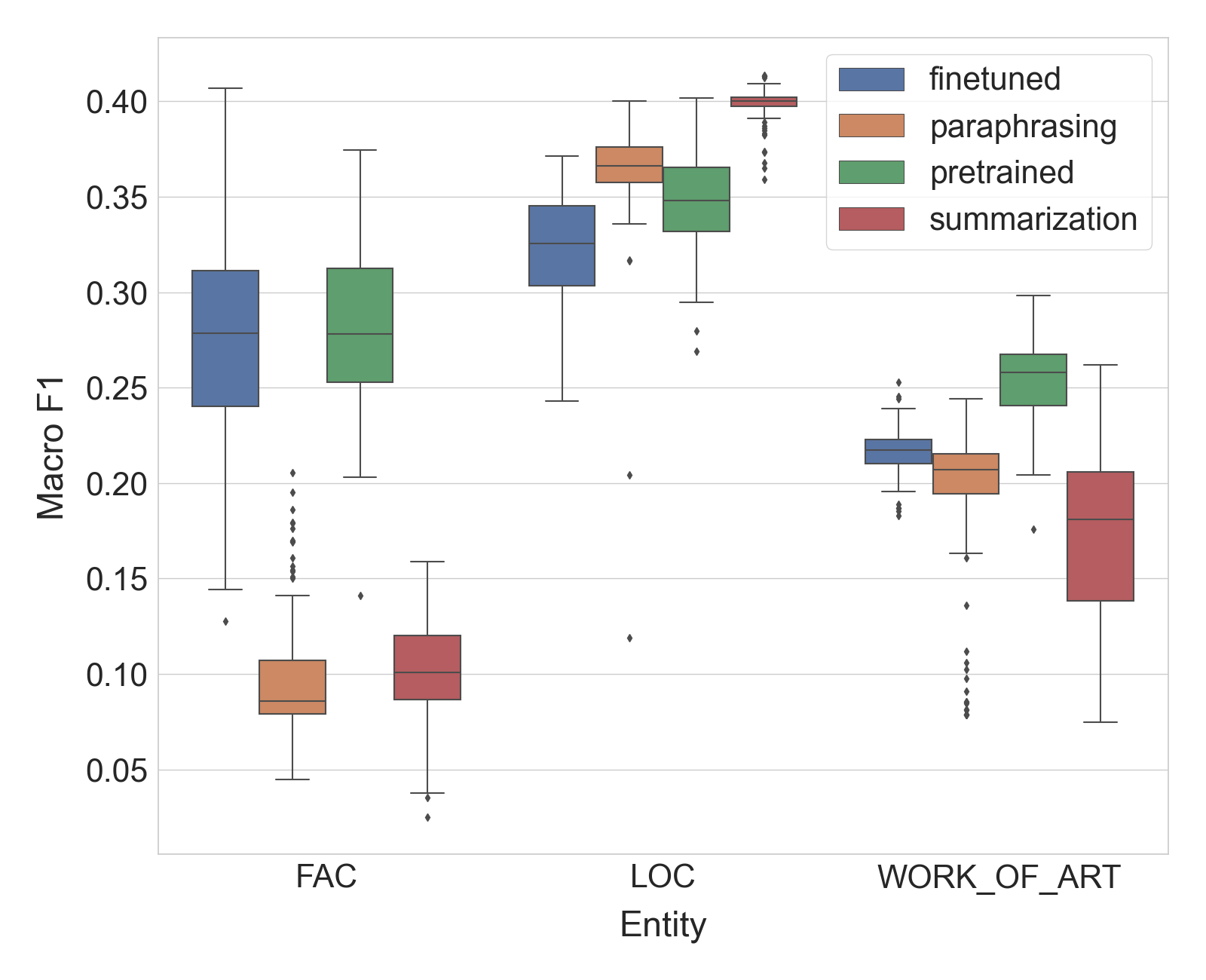}
 \caption{The figure shows the distributions of Macro F1 Score values on the test split of the OntoNotesZS dataset for each class, using the strategies described in Section \ref{sec:desc_alter_strategies} to generate 100 description variations for each class.}
 \label{fig:strategy-comparison-ontonotes}
\end{figure}

\begin{figure}[bth]
 \centering
 \includegraphics[width=\columnwidth]{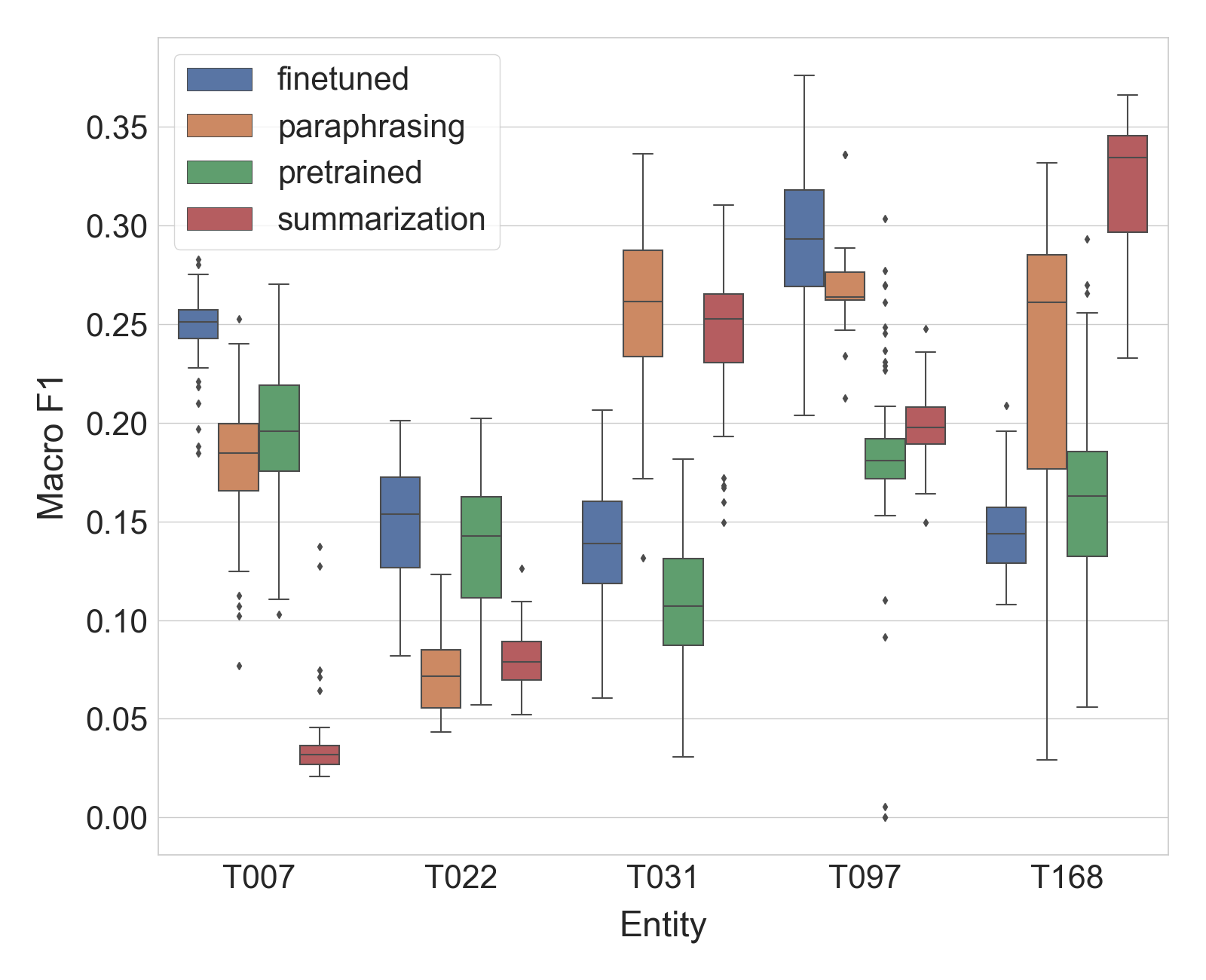}
 \caption{The figure shows the distributions of Macro F1 Score values on the test split of the MedMentions dataset for each class, using the strategies described in Section \ref{sec:desc_alter_strategies} to generate 100 description variations for each class.}
 \label{fig:strategy-comparison}
\end{figure}

\subsubsection{Relation classification}
In Table \ref{tab:ensemble_relations}, we report our evaluation of the proposed approaches on the RC task. The results we observe here are similar to what we described for entity classification where the proposed ensembling method (UDEBO) achieves a higher performance across different measures compared to the baseline ZS-BERT model that does not rely on any relation description reformulation approach. We also observe on the FewRel dataset a higher Macro F1 Score associated with most of the description enhancement variants when employed independently from each other. These results further validate the strength of the proposed approach to enhance relation descriptions employed by ZSL models to improve their performance.

\subsubsection{Descriptions enhancement strategies comparison and limitations}

Generating variations of descriptions is relatively simple, as described in Subsection \ref{sec:desc_alter_strategies}, several strategies allow to generate plausible extensions or variations of a text. Considering the results of ranking the descriptions using entropy in Section \ref{sec:experimental_results}, we analyze and discuss here the correlation between Macro F1 Score and entropy measures and the limitations of the proposed approach.

\begin{figure*}[hbt]
\center{\includegraphics[width=1\textwidth]
{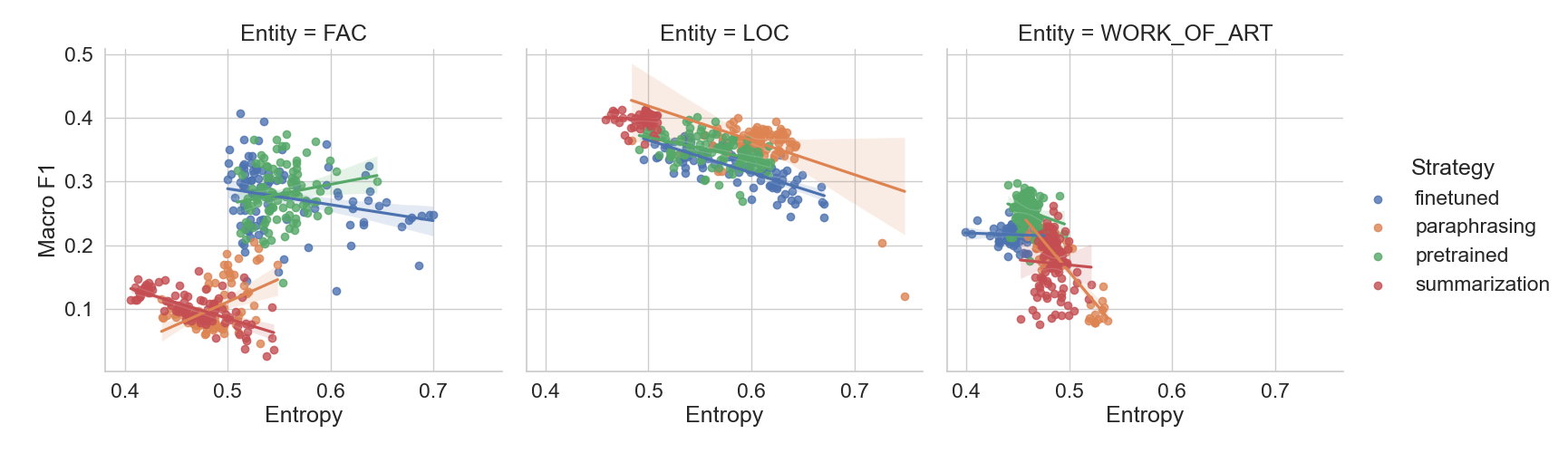}}
\caption{\label{fig:onto-notes-correlations} Analysis of the correlation between entropy and Macro F1 Score on unseen classes on the OntoNotesZS test split. Entropy can be calculated without the need for labeled data, therefore, if a correlation exists it can be used as an unsupervised heuristic to select descriptions that improve model performance.}
\end{figure*}

Figure \ref{fig:strategy-comparison-ontonotes} and Figure \ref{fig:strategy-comparison} show the distributions of the Macro F1 Score on the test split of the OntoNotesZS and the MedmentionsZS dataset for each class, using the strategies described in Section \ref{sec:desc_alter_strategies} to generate 100 description variations for each class. None of the strategies is a clear champion over all the classes. The high variance of the performance explains the fact that the ensemble method makes a better prediction as observed in Table \ref{tab:ensemble} and Table \ref{tab:ensemble_relations} thanks to successfully combining the strength of individual description alteration strategies. 
Figure \ref{fig:onto-notes-correlations} shows the correlations between Macro F1 Score and entropy for each unseen class on the OntoNotesZS test split with 100 description variations.
Although there appears to be a significant statistical correlation using a sign test with (\textit{p-value} $=0.03$) between Macro F1 Score and entropy measures on the OntoNotesZS test set, the correlation does not appear to be statistically significant in the MedMentionsZS dataset. Also, as evidenced by the results in Table \ref{tab:ensemble} and \ref{tab:ensemble_relations}, using the descriptions with minimum entropy does not seem like a good strategy for selecting descriptions. 

This phenomenon may be due to several factors like the change in the style of generated descriptions compared to the ones observed during training. Although a new description might seem more relevant, it could make the model more uncertain. See an example in Appendix \ref{appendix:example-correlations}.
The importance of this problem motivates the future study of alternative heuristics with more significant correlations, indirectly unveiling the mechanism behind zero-shot predictions.

\section{Related work} \label{sec:related}

\textbf{Zero-shot entity recognition and linking.}
Zero-shot end-to-end entity linking refers to the task of detecting and disambiguating entity mentions by linking them to an entity in a Knowledge Base (KB), without requiring new labeled data. KBs are inherently incomplete and evolve over time with the addition of new entities and relations. Zero-shot entity linking usually relies on available textual information, or other set of relations in the KB, to generalise to entity sets unseen in the training data. 

BLINK \cite{wu-etal-2020-scalable} is a BERT-based solution for Zero-shot linking of textual mentions - extracted for example using FLAIR \cite{akbik2018coling} - to entities in Wikipedia. It follows a bi-encoder architecture, each mention is encoded in a dense space, together with its context (left and right part of the input sentence). Independently, each entity in the KB is encoded in the same dense space together with its context e.g., entity description. Mentions are linked to entities in the dense space using a nearest neighbour search. To improve accuracy, candidate entities are ranked by passing each concatenated mention, its context and entity description to a more expensive cross-encoder. 

GENRE \cite{decao2021autoregressive} is a BART based model fine-tuned using a sequence to sequence objective, which claims to outperform BLINK. It is an autoregressive end-to-end entity linker, it detects and retrieves mentions and the respective entities in a KB by generating their unique textual name - left to right, token-by-token. To do so, it uses a constrained decoding strategy that forces the generated name to be in a predefined candidate set. Compared to multi-class classification models such as BLINK, GENRE has a lower memory footprint to store dense vectors for large KBs, scaling linearly with vocabulary size, not entity count, and does not need to subsample negative data during training. 

\paragraph{Zero-shot entity classification.} Entity classification consists in predicting a probability for each semantic type of an entity mention, given a set of types (e.g, organisation, organic compound). The most straightforward feature used to generalise to unseen types is the textual descriptions. For example, SMXM \cite{aly-etal-2021-leveraging} uses a cross-attention encoder to generate a vector representation for each type description and token in the input sentence and recognizes as entity types those representations that are closer to each other, including rarer classes unseen in training. It is evaluated using zero-shot adaptations of \textit{OntoNotes} \cite{pradhan-etal-2013-towards} and the domain specific biomedical dataset \textit{MedMentions} \cite{mohan2019medmentions}, it also considers \textit{out-of-KB} predictions i.e., \textit{nil} predictions for mentions that do not have a valid gold entity.

ReFinED \cite{ayoola-etal-2022-refined} is an end-to-end entity linking model optimised to perform mention detection, fine-grained entity typing (classification), and entity disambiguation in a single pass. Similar to BLINK, ReFinED uses a bi-encoder architecture modified to encode all mentions in a document simultaneously, which improves efficiency relatively to zero-shot models such as \cite{wu-etal-2020-scalable} that requires a forward-pass for each mention. Mention embeddings and entity description embeddings are projected into a shared vector space to calculate their dot product as the entity score. A fast bi-encoder combined with a score for unseen entities, computed based on the scores for entity types and description, is enough for ReFinED to obtain state-of-the-art performance on entity linking and to scale the approach from Wikipedia (5.9M entities) to Wikidata (90M entities).


The analyses in \cite{aly-etal-2021-leveraging} show that while Wikipedia descriptions work well on general entity types, they perform poorly on domain specific data, e.g. \textit{MedMentions}. They also show the impact of using annotation guidelines for descriptions to improve the transfer of knowledge from observed to unseen entities. The adoption of this approach led to a better performance compared to using a class name itself or Wikipedia passages. In particular, description vagueness, noise and negations had a negative effect, while annotation guidelines, including explicit examples and syntactic and morphological cues, improved the performance. 

\paragraph{Zero-shot relation classification.}
Textual descriptions have also been employed in the relation classification task to predict new relations that could not be observed at training time. For example, ZS-BERT \cite{DBLP:journals/corr/zs-bert} learns two functions – one to project sentences and the other to project relation descriptions into an embedding space. The objective is first to jointly minimise the distance between the embedding vectors for an input sentence and the relation description for positive entity pairs and then to classify the relation (using a softmax layer to produce a classification probability). At inference time, the prediction of unseen relation classes can be achieved through nearest neighbor search. 
Overall, using descriptions seems to improve existent zero-shot methods and expand their domains of application. Still, descriptions are not always good enough to get good predictions. Improving the accuracy of these approaches remains an open challenge. The better the separation between embedding of different relations, the more accurate the model predictions, however, as the number of unseen relations increases, it becomes more difficult to predict the right one \cite{DBLP:journals/corr/zs-bert}.

Existent ZSL methods usually rely on external knowledge from KGs, ranging from textual information, class attributes, hierarchy, domain and range constrains and relations to logic rules. There are relatively few studies evaluating their performance for unseen relations, a comparison using different external knowledge settings for zero-shot relation classification and KG completion can be seen in \cite{Geng2021BenchmarkingKZ}. To the best of our knowledge, we present the first approach to automatically predict and generate entity descriptions to improve the accuracy of entity recognition and relation classification models. 


\section{Discussion about Large Language Models}\label{sec:discussion}
In the era of Large Language Models, all kinds of problems are being solved with LLMs, that achieve outstanding results in different tasks. However, LLMs also raise some concerns, being one of the most important, the green footprint of these models. Serving a single 175 billion LLM requires at least 350 GB GPU memory using specialized infrastructure, \cite{zheng2022alpa}. This makes it unfeasible for a lot of users to use LLMs, and even if it's possible to use, there is a lot of concern about using them, specially for tasks that could be solved with smaller models. With UDEBO we try to push the research in a direction that improves the performance of small LMs to achieve results comparable to LLMs. However, to compare the performance of UDEBO against LLMs, we select 3 open-source LLMs available to the community and evaluate them. The results, discussion, and settings can be found in Appendix \ref{appendix:discussion-llms}.

\section{Conclusion and future work}\label{sec:conclusions}
In this paper, we formally defined the problem of selecting descriptions to make predictions about unseen classes in the ZSL context. To the best of our knowledge, this is the first time for entity/relation ZSL problems in which the impact of description variations on prediction performance is studied, and different methods for automatic creation of descriptions are considered. We empirically evaluated the sensitivity of two ZSL methods to description changes, and proposed 4 different strategies to enhance them using the implicit knowledge of pre-trained language models. We also studied in detail the efficacy of the proposed entropy-based heuristic to rank different description formulations, analyzing its correlation with the performance (in terms of Macro F1 Score) of the model. We observed a negative correlation between the proposed heuristic and Macro F1 Score on two out of four of the considered datasets (OntoNotesZS and FewRel). The same assumption however was not valid for the other datasets (MedMentionsZS and WikiZS), thus motivating the need to develop more effective heuristics in the future.
Finally, we described the UDEBO method, which combines the predictions obtained by the same model using different automatically generated variants of entity and relation descriptions. Our experimental results, on 4 different datasets, spanning across two different NLP tasks (Entity Classification and Relation Classification) showed how UDEBO outperforms the baselines by a significant margin and achieves new state-of-the-art results on these benchmarks under the zero-shot setting. 
Existing ensemble methods focus on ensembling different models trained on different data or models with different structures. Our work is orthogonal to these approaches, we proposed methods that consider entity/relation description variations as the hyper-parameters that need to vary. Most importantly, the description variations are not provided by the users but were generated from the initially provided descriptions. Therefore, this is a new way of creating ensembles, at least in the context of Zero-shot Entity/Relation extraction this is the first time a method for descriptions generation to diversify the pipelines and make an ensemble that improve the quality of the results is proposed.


\newpage

\section*{Limitations}

While our proposed method of boosting for zero-shot entity and relation classification shows promising results, there are several limitations that need to be acknowledged.
Firstly, ensembling methods can be computationally intensive, which can limit their applicability to large-scale datasets. Our current implementation combining multiple model predictions, by varying the descriptions, which requires a significant amount of computational resources. Therefore, future research should explore alternative methods for ensembling that are more computationally efficient.
Secondly, while we used entropy as a metric to identify helpful descriptions, it may not always be the most effective metric. Entropy measures the uncertainty or randomness of a distribution, but it may not necessarily capture the semantic relevance of a description. Therefore, there is a need for further research to develop better metrics or heuristics for identifying helpful descriptions.
Finally, we have only experimented with a limited set of generation techniques. Future research should explore ways to improve the quality and coverage of the descriptions.

\section*{Supplemental Material Statement and Reproducibility}

\paragraph{Source code availability} the source code including the frameworks for pipeline ensemble and evaluation on the standard benchmark datasets are available as an open-source github repository: \url{https://github.com/IBM/zshot}.

\paragraph{Dataset availability}
Both datasets used in our evaluation are publicly available, also included in our open-source evaluation framework. We also release the enhanced descriptions generated by the generative models used to create the ensemble pipelines. 
\bibliography{anthology,custom}



\section{Appendix}
\label{sec:appendix}

\appendix

\section{Datasets}
\label{appendix:datasets}
As mentioned in Section \ref{sec:datasets}, we evaluate our approach on four different datasets, two for EC and two for RC. For EC, we use OntoNotes \cite{pradhan-etal-2013-towards} and MedMentions \cite{mohan2019medmentions}. OntoNotes is a dataset that comprises various genres of text (news, conversational telephone speech, weblogs, usenet newsgroups, broadcast, talk shows). We use the version available in Huggingface \footnote{\url{https://huggingface.co/datasets/conll2012_ontonotesv5}} and adapt it to perform zero-shot as explained in \cite{aly-etal-2021-leveraging}, removing all the entities that are out of the split - i.e., each split has a unique set of entities, so all the entities labeled with entities out of that set are removed - removing sentences without any entity labelled and using the same train/test/dev splits, so the pre-trained model has not seen the entities in the test set neither. The entity descriptions used for OntoNotesZS (the zero-shot version of OntoNotes) were provided by the authors of \cite{aly-etal-2021-leveraging}. 

MedMentions is a corpus of Biomedical papers annotated with mentions of UMLS entities. We apply the same preprocessing steps we used for the MedMentions dataset, with the descriptions available in the official GitHub repository of \cite{aly-etal-2021-leveraging}.~\footref{footnote:smxm-repo} The version of the MedMentionsZS dataset we use is also available on Huggingface. Both of them in their zero-shot version, as proposed in \cite{aly-etal-2021-leveraging}. To convert them to the zero-shot version, we follow the following steps:
\begin{enumerate}
 \item Get the train/test/dev splits of the datasets;
 \item Collect the entities in each split;
 \item Remove entities out of the split i.e., if one entity $e$ belongs to the train split, all mentions labelled as $e$ in the test and dev splits will be replaced with the $O$ label.
 \item Remove sentences without labels. As the previous processing step ($3$) may remove all the entities of one sentence, the result dataset will have a lot of empty sentences. These sentences are removed in the final dataset.
\end{enumerate}
Table \ref{tab:medmentions_entities} and Table \ref{tab:ontonotes_entities} report the entities for each split in the dataset and the number of entities for MedMentionsZS and OntoNotesZS, respectively. As we can observe, both datasets are highly imbalanced, with some entities appearing 25007 times and some others only 89 in the case of MedMentionsZS, and 24163 and 65 times for OntoNotesZS. However, the most common entities are used only for training and the ones with fewer examples are used for validation and testing. As pointed in \cite{Xian2018}, real-world scenarios annotated data is likely to be available for the more common ones. 
\begin{table}
\centering
 \begin{tabular}{|l|l|c|}
 \hline
 Split & Entity & Count \\
 \hline
 \multirow{12}{*}{Train} &  O & 515420 \\ \cline{2-3}
  & T103 & 22360 \\ \cline{2-3}
  & T038 & 25007 \\ \cline{2-3}
  & T033 & 9824 \\ \cline{2-3}
  & T062 & 5445 \\ \cline{2-3}
  & T098 & 3574 \\ \cline{2-3}
  & T017 & 12575 \\ \cline{2-3}
  & T074 & 1165 \\ \cline{2-3}
  & T082 & 7511 \\ \cline{2-3}
  & T058 & 14779 \\ \cline{2-3}
  & T170 & 5996 \\ \cline{2-3}
  & T204 & 4922 \\ \cline{2-3}
  \cline{2-3}
  \hline
  \multirow{6}{*}{Test} &  O & 27433 \\ \cline{2-3}
  & T031 &  212 \\ \cline{2-3}
  & T097 &  360 \\ \cline{2-3}
  & T007 &  448 \\ \cline{2-3}
  & T168 &  321 \\ \cline{2-3}
  & T022 &  89 \\ \cline{2-3}
  \cline{2-3}
  \hline
  \multirow{6}{*}{Validation} &  O & 34400 \\ \cline{2-3}
  & T201 &  404 \\ \cline{2-3}
  & T091 &  196 \\ \cline{2-3}
  & T037 &  434 \\ \cline{2-3}
  & T005 &  224 \\ \cline{2-3}
  & T092 &  452 \\ \cline{2-3}
 
 \hline
 \end{tabular}
 \caption{Number of entities labelled in each split in MedMentionsZS.}
 \label{tab:medmentions_entities}
\end{table}

\begin{table}
\resizebox{\columnwidth}{!}{%
 \begin{tabular}{|l|l|r|}
 \hline
 Split & Entity & Count \\
 \hline
 \multirow{5}{*}{Train} &  O & 909142 \\ \cline{2-3}
  & ORG & 24163 \\ \cline{2-3}
  & GPE & 21938 \\ \cline{2-3}
  & DATE & 18791 \\ \cline{2-3}
  & PERSON & 22035 \\ \cline{2-3}
  \cline{2-3}
  \hline
  \multirow{4}{*}{Test} &  O & 11299 \\ \cline{2-3}
  & FAC &  149 \\ \cline{2-3}
  & LOC &  215 \\ \cline{2-3}
  & WORK\_OF\_ART &  169 \\ \cline{2-3}
  \cline{2-3}
  \hline
  \multirow{5}{*}{Validation} &  O & 36790 \\ \cline{2-3}
  & NORP &  1277 \\ \cline{2-3}
  & LAW &  65 \\ \cline{2-3}
  & EVENT &  179 \\ \cline{2-3}
  & PRODUCT &  214 \\ \cline{2-3} 
 \hline
 \end{tabular}}
 \caption{Number of entities labelled in each split in OntoNotesZS.}
 \label{tab:ontonotes_entities}
\end{table}

In Table \ref{tab:datasets_insights} we report some statistics concerning the length of sentences on both MedMentionsZS and OntoNotesZS. In both datasets, there are sentences containing only 1 token and 1 entity. The maximum number of tokens also varies across datasets and splits, with a maximum of 179 for MedMentionsZS and 210 for OntoNotesZS. 

For RC, we use the FewRel\cite{han-etal-2018-fewrel} and WikiZS \cite{sorokin2017context} datasets. FewRel is a dataset for RC compiled by collecting entity-relation triplets with sentences from Wikipedia articles, and manually filtered to ensure the data quality and class balance. We use different relations for the train and the test split to ensure the zero-shot version of the dataset. The dataset is available in the Huggingface hub.~\footnote{\url{https://huggingface.co/datasets/few_rel}} We use the \emph{train\_wiki} split in Huggingface as training split for the ZS-BERT model and the \emph{wiki\_val} as test split. Table \ref{tab:rows_entities_relations} shows the total number of sentences in FewRel, and the number of different relations for each split. There are 700 samples for each relation in each split, thus the number of sentences reported in Table \ref{tab:rows_entities_relations} is equal to the number of relations times the number of samples for each of them (e.g. train split: $44800 = 64 * 700$). Differently from FewRel, WikiZS was constructed using the Wikidata knowledge base. The dataset contains a total of 93431 sentences, each with an entity pair and a labelled relation between them. In this case, the number of instances per relation class is not balanced and we employ our own random splits containing different distinct sets of relations for the training (83 relations), validation (15 relations) and testing (15 relations) of the ZS-BERT model. More information on the dataset is contained in Table \ref{tab:rows_entities_relations}.

\begin{table*}[t!]
 \begin{tabular}{|l|l||r|r|r||r|r|r|}
 \hline
  Dataset & Split & \begin{tabular}{@{}c@{}}Mean \\ \#Tokens\end{tabular} & \begin{tabular}{@{}c@{}}Max \\ \#Tokens\end{tabular} & \begin{tabular}{@{}c@{}}Min \\ \#Tokens\end{tabular}
  & \begin{tabular}{@{}c@{}}Mean \\ \#Entities\end{tabular} & \begin{tabular}{@{}c@{}}Max \\ \#Entities\end{tabular}& \begin{tabular}{@{}c@{}}Min \\ \#Entities\end{tabular} \\
 \hline

\multirow{3}{*}{MedMentionsZS} &   train &    26 &    179 &    1 &     6 &    78 &     1 \\ \cline{2-8}
   &   test &    28 &    102 &    2 &     2 &    33 &     1 \\ \cline{2-8}
   &  validation &    28 &    119 &    4 &     2 &    12 &     1 \\ \cline{2-8}
\hline
\multirow{3}{*}{OntoNotesZS} &  train &    25 &    210 &    1 &     3 &    99 &     1 \\ \cline{2-8}
   &  test &    29 &    108 &    2 &     3 &    39 &     1 \\ \cline{2-8}
   & validation &    28 &    186 &    3 &     1 &    27 &     1 \\ \cline{2-8}
 \hline
 \end{tabular}
 \caption{Entity classification datasets details.}
 \label{tab:datasets_insights}
\end{table*}

\begin{figure*}[t!]
\center{\includegraphics[width=1.0\textwidth]
{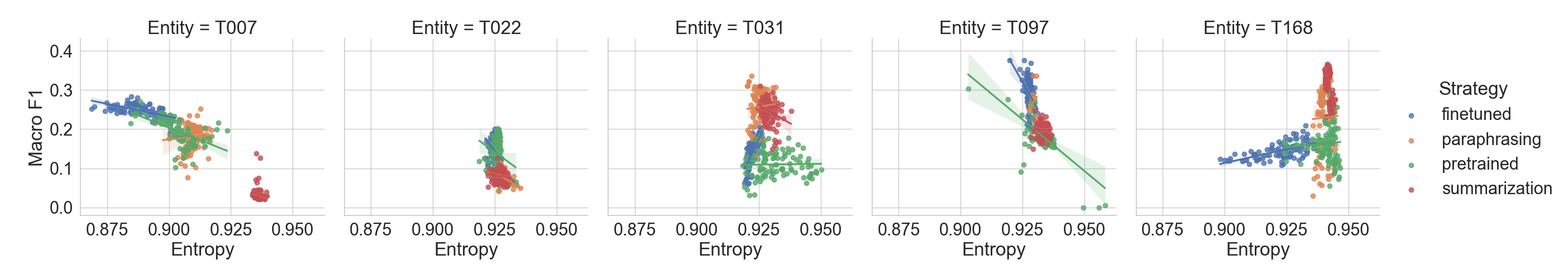}}
\caption{\label{fig:medmentions-correlations} Analysis of the correlation between entropy and Macro F1 Score on unseen classes on the MedmentionsZS test split.}
\end{figure*}

\section{Additional details on the models used for generating description variations}
\label{appendix:hp}

In this section, we report additional details on the methods used to generate description variations described in Section \ref{sec:desc_alter_strategies}.

\paragraph{Extension with pre-trained LMs.} An off-the-shelf GPT-2 pre-trained model was used for generating the variations, using the checkpoint from the Huggingface Hub.~\footnote{\url{https://huggingface.co/gpt2}}
We used $ \textit{min\_length}=80$, $ \textit{max\_length}=120 $, $ \textit{num\_beams}=8 $, $ \textit{temperature}=1 $ and $ \textit{no\_repeat\_ngram\_size}=2 $ for the generation.
 
\paragraph{Extension with a fine-tuned LM.} A model based on T5 large \cite{2020t5} and fine-tuned on the task of description generation and extension was used for generating the variations. As a starting point for the fine-tuning, the checkpoint from Huggingface Hub \footnote{\url{https://huggingface.co/t5-large}} was used. The Wikidata dataset, containing the name and the first few sentences of included Wikipedia articles where the model was fine-tuned on, was taken from Facebook Research's BLINK project.~\footnote{\url{http://dl.fbaipublicfiles.com/BLINK/entity.jsonl}}
After cleaning the data i.e., removing instances with no or too short (less than 10 words) descriptions, about 5,310,000 samples were available for training the model to perform a new sequence to sequence task using $ \textit{learning\_rate}=3e-05$ and $ \textit{epochs}=1$. The objective was to complete the input description, starting from a sub-string containing the first ten words of it. For the generation task, just the name of the description was used.
In the latter case, we set $ \textit{min\_length}=80$, $ \textit{max\_length}=120 $, $ \textit{num\_beams}=8 $, $ \textit{temperature}=1 $ and $ \textit{no\_repeat\_ngram\_size}=2 $.

\paragraph{Summarization.} A warm-started BERT2BERT (small) model fine-tuned on the CNN/Dailymail for document summarization was used for generating the descriptions variations, using the checkpoint from the Huggingface Hub.~\footnote{\url{https://huggingface.co/mrm8488/bert-small2bert-small-finetuned-cnn_daily_mail-summarization}}
We used $ \textit{min\_length}=80$, $ \textit{max\_length}=512 $, $ \textit{num\_beams}=8 $, $ \textit{temperature}=1 $ and $ \textit{no\_repeat\_ngram\_size}=2 $ for this set of experiments.

\paragraph{Paraphrasing.} A PEGASUS model fine-tuned for paraphrasing was used for generating the description variations, using the checkpoint from the Huggingface Hub.~\footnote{\url{https://huggingface.co/tuner007/pegasus_paraphrase}} 
We used $\textit{min\_length}=10$, $ \textit{max\_length}=60 $, $ \textit{num\_beams}=8 $, $ \textit{temperature}=1 $ and $ \textit{no\_repeat\_ngram\_size}=2 $ for the generation of text.

\section{Additional experiments on Entropy and Macro F1 Score correlations}
\label{appendix:add-exps-entropy-correlations}

In this section we report additional insights on the correlation analysis discussed in the paper.

\subsection{Correlations analysis of Macro F1 Score and entropy on MedmentionsZS}

Figure \ref{fig:medmentions-correlations} reports the correlations between Macro F1 Score and entropy on MedmentionsZS test-set. As discussed in the paper we did not observe any statistically significant correlations, with $\textit{p-value} = 0.50$ .

\subsection{Example of generated descriptions and entropy values}
\label{appendix:example-correlations}

\begin{table*}[!htb]
    \centering
    \fontsize{7}{7}\selectfont
    \begin{tabularx}{\linewidth}{>{\hsize=1.8cm}XX>{\hsize=1.8cm}X}
        \hline
        \arrayrulecolor{black}
        \rowcolor[rgb]{0.941,0.941,0.941} \textbf{Enhancement method} & \textbf{Description} & \textbf{Entropy}\\
        \hline
        \arrayrulecolor[rgb]{0.6,0.6,0.6}
        
        Summarization & " i\'m reporting live from right outside hospital, " says massachusetts general hospital. buildings are referred to using the name of the company or organization that uses them. names of man - made structures should be marked as fac when they refer to the structure, usually in a locative way. " he\'s reporting. here : " we\'re looking for people who don\'t know the exact location of this type " & 0.460\\
        \hline
        Summarization & " i\'m reporting live from right outside hospital, " says massachusetts general hospital. buildings are referred to as fac when they refer to the physical structure of the building itself. names of man - made structures include infrastructure ( streets, bridges ), buildings, monuments, etc. " we\'re still trying to make a difference in our lives,\'says hospital\'s director of medical examiners. & 0.538\\
        \hline
        Pretrained & Names of man-made structures: infrastructure (streets, bridges), buildings, monuments, etc. 1. The name of the building or structure is used to distinguish it from other buildings or structures. For example, a building might be called a museum or a church. A building can also be referred to as a "museum" or "church" if it is built on top of an existing structure (e.g., a university building). It is also possible to refer to a structure by its name. Examples of buildings that are known as museums include: a school building; & 0.497\\
        \hline
        Pretrained & Names of man-made structures: infrastructure (streets, bridges), buildings, monuments, etc. The number of buildings in a city is based on the size of the city and its population. The larger the number, the more likely a building is to be built. For example, if you have 10,000 people, you would expect to have a 10\% chance of having 10 buildings per square mile. However, this is not always the case. In fact, there are some buildings that are more than 10 times as large as the population of your city.  & 0.573\\
        \hline
        Paraphrasing & Buildings that are referred to using the name of a company or organization should be marked as FAC if they refer to the physical structure of the building itself. & 0.438\\
        \hline
        Paraphrasing & The names of the man-made structures should be marked FAC when references are made to the physical structure. & 0.558\\
        \hline
        Finetuned & Names of man-made structures: infrastructure (streets, bridges), buildings, monuments etc. are the names given to a number of different types of structures in the world, e.g. roads, railways and roads. The names of these structures are usually derived from the words "infrastructure" and "building" or "domestic". & 0.512\\
        \hline
        Finetuned & Names of man-made structures: infrastructure (streets, bridges), buildings, monuments etc. are the names given to the physical structures of a country or region. These names are usually derived from the name of the municipality in which they are located, or the place where it is located (e.g., city or town). For example, in the United States, the city of San Francisco is known as "San Francisco International Airport", while in Mexico, Mexico City is called "Mexico City" or "California". & 0.699\\
        \arrayrulecolor{black}
        \hline
    \end{tabularx}
    \caption{Example descriptions variations for the class FAC of OntoNotes. The worst and best variations, according to the entropy, are selected for each method.}
    \label{table:fac_descriptions_variations}
\end{table*}

\begin{table*}[!htb]
    \centering
    \fontsize{7}{7}\selectfont
    \begin{tabularx}{\linewidth}{>{\hsize=1.8cm}XX>{\hsize=1.8cm}X}
        \hline
        \arrayrulecolor{black}
        \rowcolor[rgb]{0.941,0.941,0.941} \textbf{Enhancement method} & \textbf{Description} & \textbf{Entropy}\\
        \hline
        \arrayrulecolor[rgb]{0.6,0.6,0.6}

Summarization & also included in this category are named regions such as the middle east, areas, neighborhoods, continents and regions of continents. these include mountain ranges, coasts, borders, planets, geo - coordinates, bodies of water. don't mark deictics or other non - proper nouns, but don't be marked when they are part of the location name itself. the list is based on the names of geographical locations other than gpes. & 0.465\\
\hline
Summarization & these include mountain ranges, coasts, borders, planets, geo - coordinates, bodies of water. also included in this category are named regions such as the middle east, areas, neighborhoods, continents and regions of continents. don't mark deictics or other non - proper nouns, but don't mark the names. do you know a hero? nominations are open at cnn. com / heroes. & 0.510\\
\hline

Pretrained & Names of geographical locations other than GPEs. These include mountain ranges, rivers, lakes, and streams.

For more information, please visit: & 0.492\\
\hline
Pretrained & Names of geographical locations other than GPEs. These include mountain ranges, lakes, rivers, streams, and oceans.

The following table lists the geographic locations of each of the three types of geographic data used in this report. The geographic coordinates for each type of data are listed in the table below. For more information about the geospatial data, please refer to the Geographic Information System (GIS) Web site at & 0.583\\
\hline

Paraphrasing & The geographical locations that aren't included in GPEs include mountain ranges, coasts, borders, planets, as well as the regions of continents and the Middle East. & 0.582\\
\hline
Paraphrasing & There are geographical locations beyond GPEs, including mountain ranges, coasts, borders, planets, and bodies of water. & 0.656\\
\hline

Finetuned & These are the names of geographical locations other than GPEs. These include mountain ranges, peaks, lakes, rivers, and other geographical features that are not part of a geographic area. Some of these names are also used in other places such as the United States, United Kingdom, Canada, Australia, New Zealand, France, Germany, Italy, Japan, Norway, Sweden, Switzerland and Switzerland. & 0.521\\
\hline
Finetuned & These are the names of geographical locations other than GPEs. These include mountain ranges, mountain peaks, and other geographical features, such as rivers, streams, lakes, or other bodies of water. The following is a list of names used by the United States Geological Survey (USGS) to identify geological features that are not part of the U.S. Geographic Names Information System (GNIS). & 0.618\\
\hline
        \arrayrulecolor{black}
        \hline
    \end{tabularx}
    \caption{Example descriptions variations for the class LOC of OntoNotes. The worst and best variations, according to the entropy, are selected for each method.}
    \label{table:loc_descriptions_variations}
\end{table*}

\begin{table*}[!htb]
    \centering
    \fontsize{7}{7}\selectfont
    \begin{tabularx}{\linewidth}{>{\hsize=1.8cm}XX>{\hsize=1.8cm}X}
        \hline
        \arrayrulecolor{black}
        \rowcolor[rgb]{0.941,0.941,0.941} \textbf{Enhancement method} & \textbf{Description} & \textbf{Entropy}\\
        \hline
        \arrayrulecolor[rgb]{0.6,0.6,0.6}

Summarization & the headline of the article being annotated should only be marked if they are referential. in other words, a reference to an article is markable as a work of art. paper headlines should be treated if it's referred to as an art art work. but if in the body of a text, then it is marked as art by art, but it should not be used. & 0.464\\
\hline
Summarization & paper headlines should only be marked if they are referential. in other words the headline of the article is markable as a work of art. there is a reference to an article being annotated if she is referred to. but in some words it should not be used as an art work or art, it's been used to make art more modern than it was used in the original version of this article. & 0.492\\
\hline
Pretrained & Titles of books, songs, television programs and other creations. Also available on iTunes.

Bookmark this page:  & 0.452\\
\hline
Pretrained & Titles of books, songs, television programs and other creations. Also available in English and Spanish.

In addition to the titles of titles, there are a number of other titles available for download on the Internet. The following are examples of some of the most popular titles that have been released in the past few years. For more information, please see the full list of available titles on this site. You can also check out the list at the top of this page to see which titles are currently available and which ones are not. If you have any questions about any of these titles or would like & 0.477\\
\hline

Paraphrasing & The titles of books, songs, TV programs and other creations should be marked if they are referential. & 0.471\\
\hline
Paraphrasing & It is not necessary to include quotations in the article if the headline is referential. & 0.531\\
\hline

Finetuned & The titles of books, songs, television programs and other creations. Also known as the title of a book, song, TV program, or television series, are the names of the creators of those works. Often, these titles are also referred to as "titles" or "pronunciations". In the United States, the term 'title' is used to indicate the author's title. The term is also used in other countries, such as Canada, Australia, New Zealand, and the U.S. of Azerbaijan. & 0.426\\
\hline
Finetuned & Titles of books, songs, television programs and other creations. Also referred to as "titles", are the titles given to a book, song, TV program, or other work of art. The term is derived from the Latin "titus" ("title"). It can also be used to describe the title of the work itself, the author's name, etc. & 0.454\\

        \arrayrulecolor{black}
        \hline
    \end{tabularx}
    \caption{Example descriptions variations for the class WORK\_OF\_ART of OntoNotes. The worst and best variations, according to the entropy, are selected for each method.}
    \label{table:work_of_art_descriptions_variations}
\end{table*}

\begin{table*}[!htb]
    \centering
    \fontsize{7}{7}\selectfont
    \begin{tabularx}{\linewidth}{>{\hsize=1.8cm}XX}
        \hline
        \arrayrulecolor{black}
        \rowcolor[rgb]{0.941,0.941,0.941} \textbf{Class} & \textbf{Description} \\
        \hline
        \arrayrulecolor[rgb]{0.6,0.6,0.6}
        \hline
        FAC & Names of man-made structures: infrastructure (streets, bridges), buildings, monuments, etc. belong to this type. Buildings that are referred to using the name of the company or organization that uses them should be marked as FAC when they refer to the physical structure of the building itself, usually in a locative way: "I'm reporting live from right outside [Massachusetts General Hospital]".\\
        \hline
        LOC & Names of geographical locations other than GPEs. These include mountain ranges, coasts, borders, planets, geo-coordinates, bodies of water. Also included in this category are named regions such as the Middle East, areas, neighborhoods, continents and regions of continents. Do NOT mark deictics or other non-proper nouns: here, there, everywhere, etc. As with GPEs, directional modifiers such as "southern" are only marked when they are part of the location name itself.\\
        \hline
        WORK\_OF\_ART & Titles of books, songs, television programs and other creations. Also includes awards. These are usually surrounded by quotation marks in the article (though the quotations are not included in the annotation). Newspaper headlines should only be marked if they are referential. In other words the headline of the article being annotated should not be marked but if in the body of the text here is a reference to an article, then it is markable as a work of art.\\
        \arrayrulecolor{black}
        \hline
    \end{tabularx}
    \caption{Original OntoNotes class descriptions (Source: annotation guidelines, \url{https://catalog.ldc.upenn.edu/docs/LDC2013T19/OntoNotes-Release-5.0.pdf})}
    \label{table:ontonotes_descriptions}
\end{table*}

We analyze the variations for both entity and relation descriptions.
\paragraph{Entity} Tables \ref{table:fac_descriptions_variations}, \ref{table:loc_descriptions_variations} and \ref{table:work_of_art_descriptions_variations} show some examples of the description variations generated with the different methods. The original descriptions used are shown in Table \ref{table:ontonotes_descriptions}. It can be seen that the models used for generation might generate wrong descriptions, e.g. Summarization variations in Table \ref{table:fac_descriptions_variations}, in which the model, that has been fine-tuned over CNN/Dailymail, a news dataset, replicates the style of writing as if it was a reporter. Also, some hallucinations lead the models to generate URLs, that have been removed. This means that the base models used for generating the descriptions are crucial, and they should be chosen and fine-tuned carefully for the description generation task. We leave this as a task for future work.

\paragraph{Relations} Given the relation \textbf{Film Director} described as:
\\
\newline
\textit{"director(s) of film, TV-series, stageplay, video game or similar"}.
\\
\newline
The fine-tuned approach for generating variations produces the alternative description: 
\\
\newline
\textit{The director(s) of a film, TV-series, stageplay, video game or similar is the person who directs the production of the film or television series. The term "director" is also used to describe an individual or group of people who are responsible for the creation, production, and/or directing of video games, films, television shows, or other forms of media.}.
\\
\newline
Although the generated description seems more complete and containing relevant additional information, the entropy calculated with ZS-BERT is higher in this case than when using the original description. This means that the model is more uncertain of its prediction.

\section{Discussion about Large Language Models}
\label{appendix:discussion-llms}

Recently, several large language models (LLMs) have been released demonstrating high capabilities for diverse NLP tasks including, but not limited to, text generation, question answering, text summarization and also NER and RE, \cite{workshop2023bloom}, \cite{chowdhery2022palm}. In the zero-shot setting, the LLMs perform exceptionally well. However, there are some problems that might limit the usage of LLMs for NER and RE, like the need of prompt engineering and result parsing, the token limitation, the hallucionations and out of context generation or the efficiency.
\begin{itemize}
 \item Prompt Engineering and result parsing. The performance of these models is highly dependent on the prompt used, and also the output of the model \cite{ding2021openprompt}. Thus, depending on the use case, one prompt might or might not be adequate. Also, the output has to be processed to extract the actual entities, and depending on the prompt this process can be different.
 \item Token Limitation. These models are all based on text generation, having a minimum and a maximum number of tokens to generate. Depending on these hyperparameters, the result might not be complete or might lead to hallucinations or false positives.
 \item Hallucinations and out of context generation. LLMs often suffer from hallucination and out of context generation, which in the case of NER and RE might result into entities and relations extracted that are not present in the text. Some approaches add a self-verification strategy to alleviate the hallucination issue, which requires further executions of the LLM \cite{wang2023gptner}. Moreover, common NER and RE use cases focus only on some specific entities or relations to be extracted, but these models can extract other entities and relations that may not be of interest to the user.
 \item Efficiency. Serving a single 175 billion LLM requires at least 350 GB GPU memory using specialized infrastructure, \cite{zheng2022alpa}. This makes it unfeasible for a lot of users to use LLMs, and even if it's possible to use, there is a lot of concern of the green footprint of these models.
\end{itemize}
The efficiency problem is one of the most important problems of the LLMs, and thus some approaches try to reduce the size of the models or to train new models via knowledge distillation. Recent works approach the knowledge distillation process with human rationales to improve the performance of the distilled model \cite{hsieh2023distilling}. In this approach, the human rationale adds information to the input, so the model can use it to perform the task. Similarly, the descriptions of the entities and relations are used to add information to the input. We leave to future work the usage of the descriptions for knowledge distillation. In either cases, UDEBO could be used to improve the descriptions or the human rationales to improve the performance of the models.

The focus of this work is the evaluation of the method UDEBO, and not the performance of the model itself, as a different size of the model, more pretraining, or even a different architecture could lead to changes in the results. However, we evaluate 3 LLMs, BLOOM \cite{workshop2023bloom}, FALCON \cite{almazrouei-etal-2023-falcon} and LLAMA 2 \cite{touvron2023llama}. In Table \ref{tab:llms-zs} we evaluate the largest version of the models in zero-shot. In Table \ref{tab:qlora} we fine-tune a smaller version of the models (7B) using QLORA \cite{dettmers-etal-2023-qlora}, with 0.06\%, 0.03\%, and 0.12\% trained parameters for BLOOM, FALCON, LLAMA 2, respectively.

\begin{table}[htb]
\begin{tabular}{lccc}
\hline
{Model} & {Size} & {MedMentions} & {OntoNotes} \\
\hline
BLOOM & 176B & 0.11 & 0.14 \\
FALCON & 40B & 0.12 & 0.09 \\
LLAMA 2	& 70B & 0.25 & 0.10 \\
\hline
\end{tabular}
\caption{F1-Score for LLMs evaluated on MedMentions and OntoNotes.}
\label{tab:llms-zs}
\end{table}

\begin{table}[htb]
\begin{tabular}{lccc}
\hline
{Model} & {Size} & {MedMentions} & {OntoNotes} \\
\hline
BLOOM & 7B & 0.25 & 0.00 \\ 
FALCON & 7B & 0.20 & 0.00 \\
LLAMA 2	& 7B & 0.33 & 0.09 \\
\hline
\end{tabular}
\caption{F1-Score for QLORA fine-tuned LLMs evaluated on MedMentions and OntoNotes.}
\label{tab:qlora}
\end{table}

The fine-tuned version of the models benefit in the MedMentions dataset, which is specific, but they suffer in the generic domain (OntoNotes), as they extracted entities from the training set, and not the ones in the test set. We use the following prompt (example for OntoNotes):
\newline

\noindent\fbox{\parbox{\columnwidth}{
Below is an instruction that describes a task, paired with an input that provides further context. Write a response that appropriately completes the request.
\newline

\#\#\# Instruction:
\newline

From the input context below extract instances of the following labels: [‘LOCATION’, ‘WORK\_OF\_ART’, ‘BUILDING\_NAME’]
\newline

\#\#\# Input:
}}
\end{document}